\pdfoutput=1
\documentclass[11pt]{article}

\usepackage[preprint]{acl}

\usepackage{times}
\usepackage{latexsym}

\usepackage[T1]{fontenc}

\usepackage[utf8]{inputenc}

\usepackage{microtype}

\usepackage{inconsolata}

\usepackage{graphicx}

\usepackage{amsmath}
\usepackage{booktabs}
\usepackage{multirow}
\usepackage[most]{tcolorbox}
\usepackage{tikz}
\usepackage{algorithm}
\usepackage{algorithmic}
\usepackage{pifont}
\usepackage{enumitem}

\hypersetup{colorlinks,linkcolor={red}}

\newcommand{\Model}{CyberBOT}

\title{{\Model}: Towards Reliable Cybersecurity Education via Ontology-Grounded Retrieval Augmented Generation}

\author{
\scalebox{0.95}{
 \textbf{Chengshuai Zhao\textsuperscript{\ding{171}}},
 \textbf{Riccardo De Maria\textsuperscript{\ding{171}}},
 \textbf{Tharindu Kumarage\textsuperscript{\ding{171}}},
 \textbf{Kumar Satvik Chaudhary\textsuperscript{\ding{171}}},
 } \\
\scalebox{0.95}{
 \textbf{Garima Agrawal\textsuperscript{\ding{171}}},
 \textbf{Yiwen Li\textsuperscript{\ding{170}}},
 \textbf{Jongchan Park\textsuperscript{\ding{170}}},
 \textbf{Yuli Deng\textsuperscript{\ding{171}}},
 \textbf{Ying-Chih Chen\textsuperscript{\ding{170}}},
 \textbf{Huan Liu\textsuperscript{\ding{171}}}
}
\\
 \textsuperscript{\ding{171}}School of Computing and Augmented Intelligence, Arizona State University
\\
 \textsuperscript{\ding{170}}Mary Lou Fulton Teachers College, Arizona State University
\\
    \texttt{\{czhao93,rdemari1,kskumara,kchaud13,garima.agrawal},\\
    \texttt{yiwenli2,jpark366,ydeng19,ychen495,huanliu\}@asu.edu}}

\begin{document}
\maketitle
\begin{abstract}
Advancements in large language models (LLMs) have enabled the development of intelligent educational tools that support inquiry-based learning across technical domains. In cybersecurity education, where accuracy and safety are paramount, systems must go beyond surface-level relevance to provide information that is both trustworthy and domain-appropriate. To address this challenge, we introduce {\Model}\footnote{Code:~\href{https://github.com/rccrdmr/CyberBOT}{https://github.com/rccrdmr/CyberBOT}}, a question-answering chatbot that leverages a retrieval-augmented generation (RAG) pipeline to incorporate contextual information from course-specific materials and validate responses using a domain-specific cybersecurity ontology, 
The ontology serves as a structured reasoning layer that constrains and verifies LLM-generated answers, reducing the risk of misleading or unsafe guidance. 
{\Model} has been deployed in a large graduate-level course at Arizona State University (ASU)\footnote{Video:~\href{https://youtu.be/m4ZCyS4u210}{https://youtu.be/m4ZCyS4u210}}, where more than one hundred students actively engage with the system through a dedicated web-based platform. Computational evaluations in lab environments highlight the potential capacity of {\Model}, and a forthcoming field study will evaluate its pedagogical impact.
By integrating structured domain reasoning with modern generative capabilities, {\Model} illustrates a promising direction for developing reliable and curriculum-aligned AI applications in specialized educational contexts.

\end{abstract}

\section{Introduction}
\begin{figure*}
    \centering
    \vspace{-6mm}
    \includegraphics[width=1\textwidth]{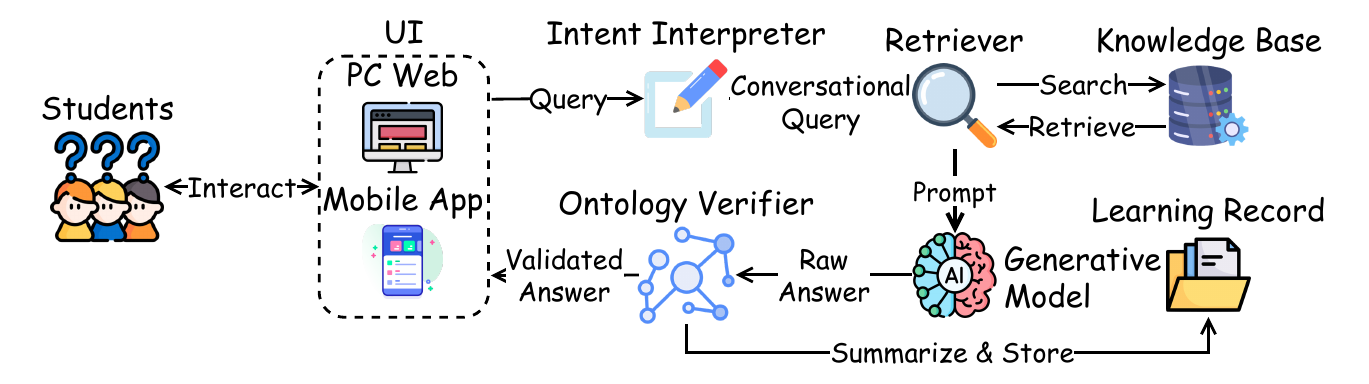}
    \vspace{-6mm}
    \caption{Framework of proposed {\Model}. Students submit queries to UI and get responses from the backend.}
    \label{fig:framework}
    \vspace{-2mm}
\end{figure*}

The integration of large language models (LLMs) into educational applications has introduced new opportunities for personalized, interactive learning experiences ~\citep{dandachi2024ai,zhao2024ontology,yekollu2024ai}. In particular, question-answering (QA) systems powered by LLMs offer the potential to support self-paced inquiry and deepen conceptual understanding~\citep{gill2024transformative,zhao2025scale}. However, despite their generative capabilities, LLMs often suffer from factual inaccuracies and hallucinations~\citep{jiang2024catching}, mainly when applied to high-stakes and technically demanding domains such as cybersecurity education~\citep{triplett2023addressing}. The risks associated with such outputs, ranging from conceptual misunderstanding to propagation of unsafe practices~\citep{tan2024glue,tan2024wolf,zou2023universal}, underscore the need for enhanced reliability in educational settings.

Retrieval-augmented generation (RAG) has emerged as a common strategy to improve response accuracy by conditioning model outputs on retrieved external documents~\citep{lewis2020retrieval}. Although this approach increases contextual relevance, it does not offer a guarantee of correctness, particularly when the retrieved context is ambiguous or incomplete~\citep{barnett2024seven,zhao2025chain}. Consequently, RAG-based systems remain susceptible to producing only loosely grounded responses in the underlying knowledge base, leading to challenges in content validity and safety.

To address these limitations, we propose {\Model}, a QA system tailored for cybersecurity education that introduces a novel ontology-based validation mechanism as a core architectural component. The system integrates a domain-specific cybersecurity ontology to assess the factual validity of LLM-generated responses. This ontology captures structured domain knowledge in the form of typed entities, relationships, and logical constraints, offering a principled framework for verifying that generated answers conform to the semantics and procedural norms of the field. Unlike traditional static knowledge bases, ontologies provide a formal representation of domain concepts and their interrelations, enabling richer and more systematic validation of model responses.

In {\Model}, the question-answering process consists of three sequential stages, each contributing to the reliability and contextual relevance of the system's output as illustrated in Figure~\ref{fig:framework}. First, an intent interpreter analyzes the multi-turn conversational history to infer the student’s underlying intent. This component reformulates the user query into a context-enriched, knowledge-intensive version, thereby enabling more effective retrieval in multi-round interactions. Second, based on the interpreted intent, relevant documents are retrieved from a curated course-specific knowledge base using retrieval-augmented generation techniques. These documents serve as the contextual foundation for generating an initial response with the LLM. Finally, the generated answer is validated using a domain-specific cybersecurity ontology, which ensures semantic alignment with authoritative knowledge and filters out hallucinated or unsafe content. This three-stage architecture allows CyberRAG to address both contextual ambiguity and factual correctness, significantly enhancing the trustworthiness and educational value of the system in real-world instructional settings.

A key strength of {\Model} lies in its real-world deployment. The system has been integrated into a live classroom setting and is accessible to more than one hundred graduate students enrolled in the spring 2025 semester of CSE 546: Cloud Computing course at Arizona State University (ASU). Students interact with the system via a dedicated web interface, submitting course-related queries and receiving validated responses to assist with their study. This deployment enables direct observation of system usage in an authentic educational environment and provides a valuable opportunity to evaluate the practical utility and pedagogical effectiveness of ontology-informed validation in question-answering systems.

Computational evaluations in lab environments highlight the potential capacity of {\Model}. To systematically assess its impact, we plan to conduct a field study at the end of the academic term. This study will examine student perceptions of answer accuracy, relevance, trustworthiness, and overall satisfaction. The findings will inform future development and offer broader insights into the role of ontology-aware validation in educational AI systems. In summary, our contributions include:

\begin{itemize}
    \renewcommand{\labelitemi}{$\star$}
    \item We propose a novel QA system {\Model} that combines RAG with cybersecurity ontologies for answer validation, reducing hallucinations, and improving factual accuracy in a specialized domain.
    \item We construct a knowledge base from class materials to ground the QA system’s responses in relevant, contextually accurate information aligned with the course curriculum.
    \item We deploy {\Model} as a user-friendly Q\&A platform, in a live classroom with more than one hundred students, with the goal of providing practical insights into the system’s performance and student engagement.
\end{itemize}

\section{The proposed \Model}
\begin{figure*}[tb]
    \centering
    \vspace{-6mm}
    \includegraphics[width=1\textwidth]{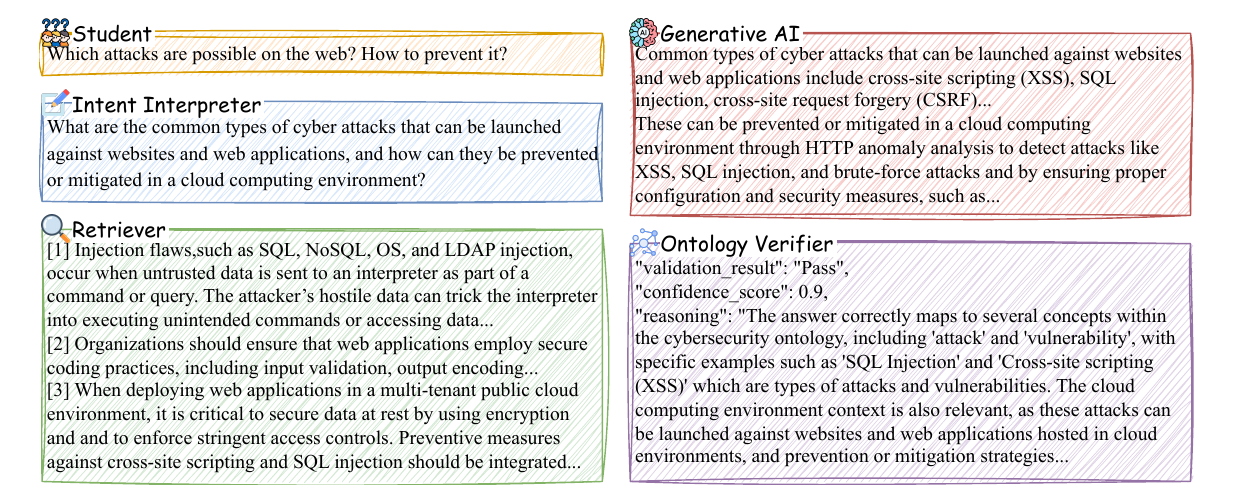}
    \vspace{-6mm}
    \caption{Illustrative of data flow in {\Model}. The response is augmented and validated in various flows.}
    \label{fig:case}
    \vspace{-2mm}
\end{figure*}

We introduce {\Model}, an ontology-aware RAG system designed for multi-turn QA in cybersecurity education. 
Broadly, the system operates in three key steps:
(I) First, an intent model interprets the student’s question based on the chat history (Section~\ref{sec:intent_model}).
(II) Next, based on the identified intent, relevant documents are retrieved from the knowledge base to augment the LLM’s response (Section~\ref{sec:retriveval_augmented_generation}).
(III) Finally, a carefully designed knowledge graph ontology is employed to validate the generated answer (Section~\ref{sec:ontology_based_validation}). The overall framework is illustrated in Figure~\ref{fig:framework}, and a step-by-step example with data flow is shown in Figure~\ref{fig:case}.

\subsection{Intent Interpreter}
\label{sec:intent_model}
Given a domain-specific question~$q$, we leverage an intent model $\mathcal{I}$ as an intent interpreter to capture the user's intention from the last~$k$-round history conservation~$c$. Then, the intent model will rewrite the current query as a knowledge-intensive conversation query~$q_c$:~$q_c=\mathcal{I}(q,c)$, which not only enriches the context for the generation and enables multi-turn retrieval. In the implementation, we design a semantic rule-based classifier to determine if a question needs to be written or not to reduce the computational cost. 

\subsection{Retrieval Augmented Generation}
\label{sec:retriveval_augmented_generation}
Based on the conversation query, the model retrieves the related document from a knowledge base. Then, the augmented context is used to prompt LLMs to generate the answer.

\subsubsection{Knowledge Base}

The knowledge base plays a critical role in supporting the responses of {\Model} and consists of two main components:
(I) A collection of common cybersecurity QA pairs curated by domain experts, derived from laboratory instruction manuals used in graduate-level advanced cybersecurity courses. These cover topics such as building intrusion detection systems and monitoring system activity.
(II) Course materials from CSE 546: Cloud Computing at ASU, including lecture slides, assignments, quizzes, and project instructions. Most of these resources are in PDF format, which we preprocess into smaller, semantically meaningful chunks before storing them in the knowledge base.

\subsubsection{Retriever}
 
For each turn in the conversation, given the conversation query $q_c$, the retriever module $\mathcal{R}$ selects the most relevant document $d$ from the corresponding course-specific knowledge base by computing similarity scores: $d = \mathcal{R}(q_c, kb)$. To accelerate retrieval, all documents are pre-encoded into vector representations, allowing for efficient similarity search within the knowledge base.

\subsubsection{Generation}
The user query, as well as the related document, is used to prompt LLM~$\mathcal{G}$ to generate preliminary answers~$a$:~$a=\mathcal{G}(q_c,d)$.

\subsection{Ontology-based Validation}
\label{sec:ontology_based_validation}

In practice, validating the generated answers is essential to prevent misinformation or misuse. To address this, we design an ontology-based validation mechanism grounded in domain knowledge. As described earlier, a knowledge graph (KG) captures factual triples in the form of entity-relationship-entity, while an ontology defines high-level domain concepts and their semantic relationships in a structured hierarchy.

Our validation process begins by extracting and analyzing key concepts and their relationships from the course materials. These are then distilled by cybersecurity experts into a domain-specific ontology $o$, encapsulating essential patterns and logical structures. Finally, we employ an ontology verifier~$\mathcal{V}$ to assess whether the generated answer aligns with the ontology: $r = \mathcal{O}(q_c, a, o)$, where $r \in (0, 1)$ represents the validation score. Answers falling below a certain threshold are flagged and rejected to ensure the reliability and domain-appropriateness of the system’s responses.

\subsection{Student Learning History}
To facilitate personalized learning, we designed a user management system, which tracks each user's learning log and stores it in the backend while anonymizing personal information. 

\begin{figure*}[tb]
    \centering
    \vspace{-6mm}
    \includegraphics[width=1\textwidth]{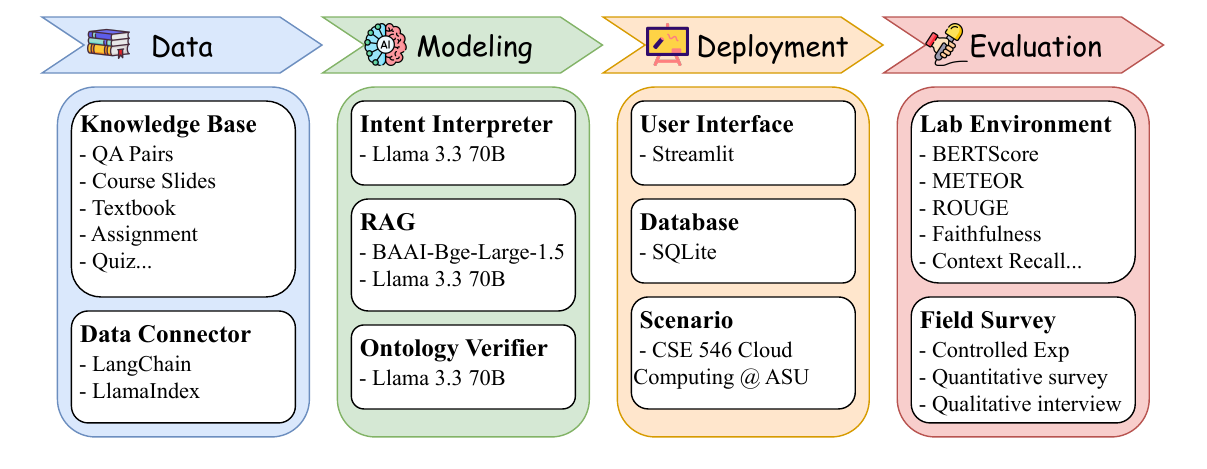}
    \vspace{-6mm}
    \caption{Pipeline of the project. The key details for each step are elaborated.}
    \label{fig:pipeline}
    \vspace{-2mm}
\end{figure*}

\section{System Deployment}
Our system architecture, as shown in Figure~\ref{fig:pipeline}, is designed to streamline the entire workflow from data ingestion to end-user interaction and evaluation. Here we discuss the deployment stage, detailing the user-facing and backend components.

\subsection{User Interface}
We build a simple, web-based front-end using Streamlit, an open-source Python framework that allows rapid development of interactive web apps. Users enter queries or select tasks, and the interface displays the system’s responses in real-time.

\subsection{Backend}
All domain-specific materials (e.g., QA pairs, slides, assignments) are stored in a unified repository. Texts are split into chunks with 512 tokens to facilitate efficient retrieval.
Each chunk is embedded using BAAI-Bge-Large-1.5 and stored in a FAISS index. User queries are similarly embedded, and nearest-neighbor search identifies the most relevant chunks.
A Llama 3.3 70B model classifies queries according to task-specific intentions. Based on these results, the system retrieves top-3 relevant chunks from FAISS for context.
Another LLlama 3.3 70B model fuses the retrieved context and the user query to generate a coherent response. This step handles reasoning and synthesizes domain knowledge for the final output.
The Ontology Verifier, powered by Llama 3.3 70B, evaluates the generated responses by verifying their alignment with our domain-specific ontology. A lightweight SQLite database is used to store session data and interaction logs for subsequent analysis.

\subsection{Device and Hardware}
We encapsulate the {\Model} into a Docker environment and develop the system in a dedicated server with an A100 80GB. For convenience, we use the Together AI API to enable all the embedding models or LLMs. For security reasons, a VPN with ASU credentials will be needed to access the system outside the university network.

\section{Application}
We have deployed {\Model} into CSE 546: Cloud Computing for the 2025 Spring semester. The application scenario shows retrieval augmented generation, and learning history track (Section~\ref{sec:use_scenario}). We conduct comprehensive lab experiments to evaluate the effectiveness of {\Model} (Section~\ref{sec:computational_evaluation}). Furthermore, we design controlled experiments and plan to conduct field surveys to evaluate the tool within an educational context~(section~\ref{sec:educational_evaluation}).

\subsection{Use-Case Illustration}
\label{sec:use_scenario}

Users can input domain-specific questions through the dialogue interface. The RAG system retrieves relevant documents from the appropriate knowledge base and generates ontology-validated responses. If a query falls outside the scope of the defined domain ontology, the response is rejected to maintain reliability and relevance. Given the highly specialized nature of the knowledge base, the system serves as an effective AI assistant for:
(I) learning cybersecurity and cloud computing concepts grounded in the course material,
(II) assisting with assignment-related queries by providing step-by-step explanations, and
(III) offering fine-grained, hands-on guidance for course projects, such as code completion and framework design. Additionally, all interactions, including user queries and corresponding responses, are stored under each user's account. These historical conversation logs can serve as a foundation for developing individualized learning paths and enabling personalized educational experiences in the future.

\subsection{Computational Evaluation}
\label{sec:computational_evaluation}

\paragraph{Dataset.} To evaluate the performance of {\Model} in a controlled lab setting, we use CyberQ~\citep{agrawal2024cyberq}, an open-source dataset comprising approx 3,500 open-ended cybersecurity QA pairs across topics such as tool usage, setup instructions, attack analysis, and defense techniques. The dataset includes questions of varying complexity, categorized into Zero-shot (1,027), Few-shot (332), and Ontology-Driven (2,171) types, making it well-suited for testing multi-level question answering.

\paragraph{Metric.} We consider two categories of metrics: (I) QA-based metrics, which evaluate the quality of generated answers, including BERTScore~\citep{zhangbertscore}, METEOR~\citep{banerjee2005meteor}, ROUGE-1~\citep{lin2004rouge}, and ROUGE-2~\citep{lin2004rouge}. (II) RAG-based metrics, which assess the retrieval effectiveness and accuracy covering Faithfulness, Answer Relevancy, Context Precision, Context Recall, and Context Entity Recall. We implement RAG-based metric using RAGAS~\citep{es2024ragas}.

\begin{table}[th]
\centering
\resizebox{\linewidth}{!}{
\begin{tabular}{lcccc}
\toprule
\textbf{Metric} & \textbf{ZS} & \textbf{FS} & \textbf{OD} & \textbf{AVG} \\
\midrule
\multicolumn{5}{l}{\textbf{QA-based Metrics}} \\
BERTScore~$\uparrow$        & 0.929 & 0.946 & 0.933 & 0.933 \\
METEOR~$\uparrow$           & 0.786 & 0.859 & 0.786 & 0.793 \\
ROUGE-1~$\uparrow$          & 0.649 & 0.788 & 0.641 & 0.657 \\
ROUGE-2~$\uparrow$          & 0.598 & 0.720 & 0.593 & 0.606 \\
\midrule
\multicolumn{5}{l}{\textbf{RAG-based Metrics}} \\
Faithfulness~$\uparrow$         & 0.813 & 0.891 & 0.760 & 0.788 \\
Answer Relevancy~$\uparrow$     & 0.983 & 0.986 & 0.983 & 0.983 \\
Context Precision~$\uparrow$    & 0.989 & 1.000 & 0.996 & 0.994 \\
Context Recall~$\uparrow$       & 0.991 & 0.997 & 0.995 & 0.994 \\
Context Entity Recall~$\uparrow$& 0.939 & 0.951 & 0.967 & 0.957 \\
\bottomrule
\end{tabular}}
\caption{Performance of {\Model} for QA-based and RAG-based metrics across various datasets.}
\vspace{-3mm}
\label{tab:performance}
\end{table}

\paragraph{Main result.} We summarize the main result in Table~\ref{tab:performance}. (I) Generally, the proposed tool achieves satisfaction across both QA-based metrics and RAG-based metrics, e.g., {\Model} achieves an average of BERTScore and Context Recall of 0.933 and 0.994, respectively, which indicates the system not only generates high-quality answers but it also can retrieve the very relevant document from the knowledge base. (II) The framework produces higher scores in the FS category than those in ZS and OD, which may be because the QA pairs in the FS leverage in-context learning examples and are thus more consistent. (III) Among the QA-based metrics, the system achieves superior performance under BERTScore compared to others. It suggests that the generated answer has good overall semantic similarity while producing various words and paraphrases. (IV) Among the RAG-based metrics, we can observe that the results under Answer Relevancy, Context Precision, and Context Recall are very competitive, which showcases the framework benefits from related documents as references. However, the Faithfulness score is slightly lower because the model will leverage its own knowledge to generate answers when there is no closely relevant material in the knowledge base.

\subsection{Educational Impact Evaluation}
\label{sec:educational_evaluation}
\paragraph{Controlled experiment.} To evaluate the effectiveness of {\Model}, a domain-specific chatbot for CSE 546: Cloud Computing coursework, we conduct a quasi-experimental study involving 77 computer science ASU graduate students. Using Monte Carlo random assignment \citep{metropolis1949monte} stratified by gender, 39 students gain chatbot access (experimental group), while 38 students complete coursework without AI support (control group). The chatbot, built specifically on course materials (e.g., textbooks, slides, project instructions), is the only permitted AI assistance. External AI tools are explicitly prohibited. We adopt a mixed-method approach using both quantitative and qualitative data analyses.

\paragraph{Quantitative analysis.} Quantitative data analyzed included student learning outcomes (e.g., quizzes, projects, summative tests) and three waves of surveys. A pre-survey measured baseline cognitive load (i.e., intrinsic, extraneous, and germane load)~\citep{leppink2013development}, initial AI literacy (i.e., awareness, usage, evaluation, ethics)~\citep{wang2023measuring}, and collected demographic information and prior academic performance. A detailed illustration can be found in Appendix~\ref{app:education_evaluation_metric}. After course project 1, the first post-survey reassessed cognitive load and AI literacy in both groups, while capturing chatbot usage frequency, perceived usefulness, and patterns of interaction for the experimental group. The control group's posttest focused instead on traditional learning resources utilized (e.g., textbooks, notes, peers). The second post-survey, conducted after Project 2, followed the same structure to assess sustained impacts over time. The complete surveys are elaborated in Appendix~\ref{app:survey}.

\paragraph{Qualitative analysis.} Follow-up semi-structured interviews are conducted with fifteen purposefully sampled chatbot users, representing varied usage levels and perceived usefulness, as well as undergraduate and graduate perspectives. Interviews explore chatbot interaction patterns, perceived impacts on cognitive load, learning strategies, and students’ broader attitudes toward AI in education. Qualitative data will be analyzed via thematic coding~\citep{braun2006using} to enrich and contextualize survey findings.

\paragraph{Data analysis.} For instrument reliability, we used previously validated scales: the Cognitive Load scale (Cronbach’s $\alpha$ = .82–.92~\citep{leppink2013development}, and the Artificial Intelligence Literacy Scale (AILS, Cronbach’s $\alpha$ = .76–.87~\citep{wang2023measuring}. Data analysis includes an analysis of covariance (ANCOVA) to compare student learning outcomes between groups, controlling for pre-survey scores, and a multiple regression analysis to examine how cognitive load, AI literacy, and chatbot usage predict learning outcomes.
 
\subsubsection{Impact for Education}
These quantitative outcomes and qualitative insights offer practical implications for designing AI-supported educational interventions.  Unlike general-purpose AI tools, {\Model}'s close alignment with course-specific learning objectives potentially reduces students' extraneous load and increases germane load ~\citep{sweller1988cognitive}, enabling deeper conceptual engagement and improved problem-solving. The evaluation draws insights from the Technology Acceptance Model~\citep{davis1989technology}, highlighting the importance of students' perceived usefulness and frequency of chatbot use for successful technology integration.

\section{Related Work}

\paragraph{AI systems for cybersecurity education.}

Recent studies have emphasized the importance of AI-driven tools in facilitating inquiry-based learning in cybersecurity education~\citep{grover2023cybersecurity,wei2023cybersecurity,ferrari2024cybersecurity}. Among such tools, knowledge graphs and ontologies have been used to structure domain knowledge and support educational applications~\citep{deng2021problem,deng2021neocyberkg}. Agrawal et al.~\citep{agrawal2023aiseckg, agrawal2022building} introduced AISecKG, a cybersecurity ontology designed to support intelligent tutoring and educational knowledge modeling. Subsequently, the CyberQ dataset~\citep{agrawal2024cyberq} was constructed using AISecKG to generate high-quality QA pairs via Ontology-based LLM prompting.

\paragraph{Domain-specific retrieval-augmented generation.}

In educational settings, RAG enables systems to provide contextually relevant and curriculum-aligned responses~\cite {dakshit2024faculty,liu2024hita,modran2024llm}. For instance, RAG has retrieved textbook sections or course notes to support complex student queries~\citep{alawwad2025enhancing, castleman2023examining}. Despite these advances, when retrieval fails to provide comprehensive or unambiguous context, LLMs may still hallucinate facts or generate inconsistent answers~\citep{elmessiry2024navigating,li2024grammar, DBLP:journals/corr/abs-2407-12216}.

\paragraph{Ontology-grounded answer validation.}

Prior work has explored using knowledge graphs and ontologies to improve consistency~\citep{hussien2024rag,de2024integrating}, with ontologies modeling domain-specific rules for verifying generated responses~\citep{majeed2025type}. While ontology-based methods have shown promise in tasks like automated grading and question generation~\citep{majeed2025type}, they are often used statically and not integrated into the response generation process.

\section{Conclusion}

We present {\Model}, an ontology-grounded RAG assistant designed to support reliable and context-aware cybersecurity education. {\Model} leverages an intent interpreter to capture the educational dialogue context and employs an ontology verifier to ground generated responses in domain-specific rules and constraints. The system has been deployed in CSE 546: Cloud Computing, serving over one hundred graduate students. Comprehensive evaluations in a controlled lab setting, along with preliminary field surveys, highlight {\Model}’s reliability and effectiveness. This work not only demonstrates the potential of specialized NLP systems in education but also opens new avenues for advancing ontology-guided approaches in cybersecurity learning. Future work will focus on enhancing the system with personalized learning capabilities tailored to individual users.

\section*{Limitations}
While {\Model} demonstrates promise in improving factual grounding for cybersecurity education, several limitations merit discussion. (I) First, the system’s accuracy depends on the quality and coverage of its curated knowledge base and ontology. Incomplete or outdated resources could lead to gaps in domain coverage, especially as cybersecurity threats and best practices evolve rapidly. (II) Second, the current deployment focuses on a single graduate-level course with a limited sample size; thus, findings may not generalize to diverse educational settings or other technical domains. (III) Third, the ontology-based validation primarily checks compliance with known concepts and relationships, leaving truly novel or emergent cybersecurity issues outside its purview. (IV) Finally, the computational overhead of real-time retrieval and validation, especially for large-scale student cohorts, poses practical challenges for broader adoption.  Future work should address these gaps by continuously updating the ontology, exploring more robust approaches for out-of-scope queries, and developing resource-efficient deployment solutions.

\section*{Ethical Considerations}
Our deployment of {\Model} for cybersecurity education carries both pedagogical benefits and important ethical responsibilities. (I) Firstly, privacy and data protection are paramount: although we store user queries and responses to facilitate personalized learning, all identifying information is anonymized and handled in accordance with institutional privacy guidelines. (II) Secondly, informed consent is integral to data collection; students are clearly notified about data usage and have the option to opt out of analytics where feasible. (III) Thirdly, bias and fairness must be considered, as large language models can inadvertently reinforce stereotypes or produce biased content. We employ continuous monitoring and prompt engineering to minimize such risks, though eliminating them entirely remains challenging. (IV) Fourthly, misuse prevention is essential in a high-stakes domain like cybersecurity. While {\Model} focuses on defending against threats and promoting safe practices, it could inadvertently reveal vulnerabilities or unsafe tips if misapplied. We mitigate these risks through ontology-based validation and guardrails that flag or reject potentially harmful or misleading outputs. (V) Finally, academic integrity is upheld by maintaining clear course policies on AI-assisted work. The proposed {\Model} system is intended to supplement, not replace, student effort.

\bibliography{custom}
\clearpage

\appendix

\section{Details for Human Evaluation}
\subsection{Explanation of Education Metric}
\label{app:education_evaluation_metric}
The evaluation metrics for this study incorporate two robust and validated scales: the Cognitive Load Questionnaire~\citep{leppink2013development} and the Artificial Intelligence Literacy Scale (AILS)~\citep{sweller1988cognitive, leppink2013development}, which captures three distinct dimensions: intrinsic, extraneous, and germane load. Intrinsic load pertains to the inherent complexity of the course content as perceived by the learners, influenced by their existing knowledge~\citep{leppink2013development}. Extraneous load reflects the cognitive effort imposed by instructional features that do not directly facilitate learning, often due to unclear or ineffective presentation~\citep{leppink2013development}. Germane load represents the cognitive resources devoted to meaningful learning activities, contributing to deeper understanding and schema acquisition~\citep{leppink2013development}.

The AI Literacy Scale assesses students' competence and comfort in interacting with AI technologies through four subconstructs: awareness, usage, evaluation, and ethics~\citep{wang2023measuring}. Awareness involves recognizing and understanding AI technologies, while usage measures the practical ability to effectively operate AI applications~\citep{wang2023measuring}. Evaluation assesses the critical capacity to analyze AI tools and their outcomes, and ethics gauges the awareness of ethical responsibilities, privacy, and risks associated with AI use~\citep{wang2023measuring}.

These metrics together provide a comprehensive evaluation framework, enabling analysis of how CyberBOT impacts both cognitive load and AI literacy among students, thereby offering insights into its effectiveness in enhancing learning outcomes within a STEM-focused educational context.

\subsection{Illustrative of Survey}
\label{app:survey}

To ensure transparency and facilitate replicability, we provide below the public links to the posttest surveys used in our quasi-experimental design. Participants in the \emph{experiment group} completed two surveys two specific course milestones, while a \emph{control group} completed parallel versions. Each survey collects data on cognitive load, AI literacy, and various measures of user experience or reliance on conventional resources. All personal or identifying data have been removed or masked in these public versions to protect participant privacy. The survey for each group at each project is provided:

\begin{itemize}[itemsep=0pt]
    \item Posttest 1 (Experiment Group): \href{https://docs.google.com/forms/d/e/1FAIpQLSd7R2rLd1wW1N-f2b95b4LM0Pjghc05hO8WkpnBkCBSXXsIYg/viewform?usp=dialog}{Link}
    \item Posttest 1 (Control Group): \href{https://docs.google.com/forms/d/e/1FAIpQLSf0xGrQXdYI_WNXsVjJlOlLR7kza9ZdoCTfvHcBKP-QjnQdzQ/viewform?usp=dialog}{Link}
    \item Posttest 2 (Experiment Group): \href{https://docs.google.com/forms/d/e/1FAIpQLSfoWFfdopFQ5uE7ZUe2mdUJ6q_Hs94dnUvFBHU99nIwdVpEKw/viewform?usp=dialog}{Link}
    \item Posttest 2 (Control Group): \href{https://docs.google.com/forms/d/e/1FAIpQLSfguYi4D9fo-sV2hDr975L997OVQFrUt-iYL1xE2LaNWGUorA/viewform?usp=dialog}{Link}
\end{itemize}

These instruments, adapted from established scales on cognitive load \citep{leppink2013development} and AI literacy \citep{wang2023measuring}, also include course-specific items that gauge student perspectives on {\Model} usage or standard (non-AI) learning resources. Aggregated results and analyses will be shared in a future publication to further illuminate the system’s pedagogical impact.

\subsection{Qualitative Interview}
\label{app:interview}
We design the following interview questions to qualitatively evaluate {\Model}:
\begin{itemize}
    \item Can you walk me through how you used {\Model} while working on course Project 1 or Project 2?
    \item What kinds of questions did you typically ask the chatbot? What were you hoping to get out of those interactions?
    \item Did the chatbot's responses usually help you? Can you recall a time when it was especially helpful, or not very helpful?
    \item Did using the chatbot change how confident you felt while working on the project? Why or why not?
    \item How did using the chatbot affect how mentally demanding the project felt?
    \item Compared to other tools or resources (like lecture notes, classmates, or online forums), how did {\Model} fit into your learning strategy?
    \item What do you think are the strengths and limitations of using a course-trained chatbot for learning?
    \item Do you have any concerns about using AI tools like this in a learning environment? 
\end{itemize}

\section{System Screenshot and Examples}
\begin{figure*}[th]
    \centering
    \includegraphics[width=\textwidth]{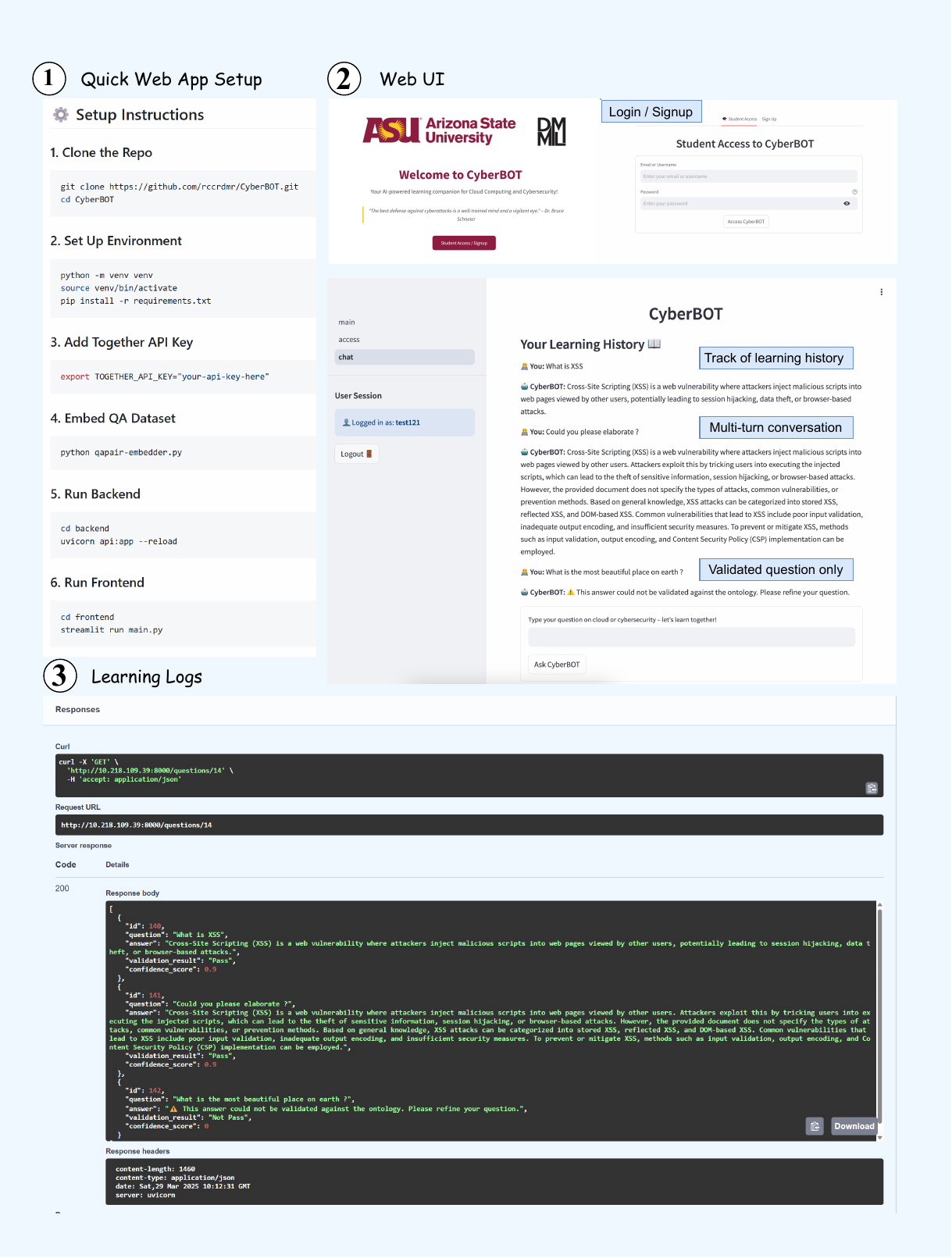}
    \caption{A live running example of {\Model} system. (1) The quick web app Setup section provides a complete step-by-step deployment guide, including cloning the repository, setting up the environment, configuring the API key, and running the backend and frontend services. (2) The web UI section showcases the login/signup interface and user dashboard, allowing personalized access and interaction. (3) The system records and displays the user's learning history, including follow-up questions and ontology-based answer validation. (4) The learning logs demonstrate backend validation of QA pairs with pass/fail results, confidence scores, and ontology-aligned reasoning
}
    \label{fig:screenshot}
\end{figure*}

Below, we present a live instance of our {\Model} system, showcasing its complete setup, user interface, and representative learning logs, as illustrated in Figure~\ref{fig:screenshot}. This example highlights the end-to-end workflow and how the proposed system supports real-time interaction for cybersecurity learning.

\clearpage
\onecolumn
\section{Illustrative of Prompt}
\label{app:prompt}
We provide here the core prompts that guide {\Model}'s functionality, including prompts of the intent interpreter, the LLM, and the ontology verifier. These prompts serve as the backbone for transforming raw user queries into validated, domain-specific responses within the cybersecurity context.

\subsection{Prompt of Intent Interpreter}
Before the system retrieves relevant documents, the intent interpreter prompt helps discern the user’s underlying goals or question types. By examining the conversation history, this prompt rewrites or augments queries to align them with the structured format required for multi-round retrieval. Below, we present the specific instructions that the intent interpreter relies upon to carry out this task.
\tcbset{
    promptstyle/.style={
        colback=gray!5,
        colframe=black,
        fonttitle=\bfseries,
        boxrule=0.5mm,
        sharp corners,
        enhanced,
        breakable,
        width=1\linewidth
    }
}
\begin{tcolorbox}[promptstyle, title=Intent Interpreter Prompt]
\small
\ttfamily
You are an assistant that rewrites vague or follow-up user questions based on previous conversation history.
Given the chat history and current question, rewrite the question to make it fully self-contained, specific, and intent-aware.

\begin{verbatim}
CHAT HISTORY:
{memory}

CURRENT QUESTION:
{current_question}

REWRITTEN QUESTION:
\end{verbatim}
\end{tcolorbox}

\subsection{Prompt of Large Language Model}
After the user’s intent is clarified and relevant contextual documents are retrieved, the large language model prompt orchestrates the actual generation of answers. It fuses the rewritten query, selected context, and any additional instructions to produce coherent, domain-focused responses. The following excerpt outlines the instructions that drive our LLM-based generation process.
\begin{tcolorbox}[promptstyle, title=Large Language Model Prompt]
\small
\ttfamily
\begin{verbatim}
DOCUMENT:
document

QUESTION:
question

INSTRUCTIONS:
Answer the user's QUESTION using the DOCUMENT text above.
Keep your answer grounded in the facts of the DOCUMENT.
If the DOCUMENT does not contain the facts to answer the QUESTION, 
give a response based on your knowledge.

Answer concisely and factually without extra commentary:
\end{verbatim}
\end{tcolorbox}

\subsection{Prompt of Ontology Verifier}
Once the model yields a tentative answer, the ontology verifier prompt is responsible for checking whether the response adheres to the cybersecurity-specific ontology. This involves confirming that crucial domain rules, relations, and constraints are not violated. Below is an illustration of how we guide the ontology verifier to perform its validation.
\begin{tcolorbox}[promptstyle, title=Ontology Verifier Prompt]
\small
\ttfamily
\begin{verbatim}
Your task is to evaluate whether the ANSWER correctly aligns with the ONTOLOGY provided below.

Return ONLY a JSON response in the format:
{
"validation_result": "Pass" or "Not Pass",
"confidence_score": CONFIDENCE_SCORE_HERE (between 0 and 1),
"reasoning": "A brief explanation of why the answer is valid or not."
}

DO NOT include anything outside of this JSON structure.

Here are a few examples:

---
Example 1 (Cybersecurity - Valid Answer, High Confidence):
QUESTION: What is a vulnerability in cybersecurity?
ANSWER: A vulnerability is a weakness in a system that can be exploited by an attacker.
EXPECTED VALIDATION RESPONSE:
{
"validation_result": "Pass",
"confidence_score": 0.95,
"reasoning": "Answer maps to 'system, can_expose, vulnerability' and 'attacker, can_exploit,
vulnerability'."
}

---
Example 2 (Cloud Computing - Valid Answer, High Confidence):
QUESTION: What is virtualization in cloud computing?
ANSWER: Virtualization is a technique that allows multiple virtual machines to run on a single
physical system.
EXPECTED VALIDATION RESPONSE:
{
"validation_result": "Pass",
"confidence_score": 0.92,
"reasoning": "Answer maps to 'Concept/technique = Virtualization' in cloud computing ontology."
}

---
Example 3 (Cybersecurity - Valid Answer, Medium-High Confidence):
QUESTION: What tool can be used to analyze vulnerabilities?
ANSWER: A logging tool.
EXPECTED VALIDATION RESPONSE:
{
"validation_result": "Pass",
"confidence_score": 0.68,
"reasoning": "Although brief, the answer is grounded in concepts like 'tool' and 'can_analyze
vulnerability'."
}

---
Example 4 (Cloud Computing - Valid Answer, Medium-High Confidence):
QUESTION: What techniques are used for load distribution in cloud computing?
ANSWER: Load balancing and auto-scaling are common techniques.
EXPECTED VALIDATION RESPONSE:
{
"validation_result": "Pass",
"confidence_score": 0.7,
"reasoning": "Answer correctly reflects cloud computing techniques from the ontology."
}

---
Example 5 (Cybersecurity - Vague Answer, Low Confidence):
QUESTION: What are security techniques in cybersecurity?
ANSWER: Techniques are used to protect systems.
EXPECTED VALIDATION RESPONSE:
{
"validation_result": "Not Pass",
"confidence_score": 0.4,
"reasoning": "Answer is too vague and not grounded in specific ontology concepts like
'Risk Assessment' or 'HoneyPot'."
}

---
Example 6 (Cloud Computing - Vague Answer, Low Confidence):
QUESTION: What are characteristics of cloud computing?
ANSWER: Cloud computing has many features.
EXPECTED VALIDATION RESPONSE:
{
"validation_result": "Not Pass",
"confidence_score": 0.35,
"reasoning": "Answer is too vague and does not mention ontology-grounded concepts like
'on-demand self-service' or 'resource pooling'."
}

---
Example 7 (Neither - Irrelevant Answer, Zero Confidence):
QUESTION: What is the capital of France?
ANSWER: Paris is the capital of France.
EXPECTED VALIDATION RESPONSE:
{
"validation_result": "Not Pass",
"confidence_score": 0.0,
"reasoning": "Answer is factually correct but completely unrelated to cybersecurity or
cloud computing ontology."
}

Now evaluate the actual input below:

QUESTION:
{question}

ANSWER:
{answer}

ONTOLOGY:
{ontology_text}
\end{verbatim}
\end{tcolorbox}

\end{document}